\documentclass[sn-mathphys-num,iicol]{sn-jnl}

\usepackage{graphicx}%
\usepackage{multirow}%
\usepackage{amsmath,amssymb,amsfonts}%
\usepackage{amsthm}%
\usepackage{mathrsfs}%
\usepackage[title]{appendix}%
\usepackage{xcolor}%
\usepackage{textcomp}%
\usepackage{manyfoot}%
\usepackage{booktabs}%
\usepackage{algorithm}%
\usepackage{algorithmicx}%
\usepackage{algpseudocode}%
\usepackage{listings}%
\usepackage{graphicx}
\usepackage{tikz}
\usepackage{hyperref} 
\usepackage{multicol}

\theoremstyle{thmstyleone}%
\theoremstyle{thmstyletwo}%

\theoremstyle{thmstylethree}%

\begin{document}

\title[Article Title]{ReCellTy: Domain-Specific Knowledge Graph Retrieval-Augmented LLMs Reasoning Workflow for Single-Cell Annotation}




\author[1]{\fnm{Dezheng} \sur{Han}}\email{dezhenghan@mail.sdu.edu.cn}
\equalcont{These authors contributed equally to this work.}

\author[2]{\fnm{Yibin} \sur{Jia}}\email{jiayibin@email.sdu.edu.cn}
\equalcont{These authors contributed equally to this work.}

\author[3,4]{\fnm{Ruxiao} \sur{Chen}}\email{rchen117@jh.edu}

\author[1]{\fnm{Wenjie} \sur{Han}}\email{wenjiehan@mail.sdu.edu.cn}

\author*[1]{\fnm{Shuaishuai} \sur{Guo}}\email{shuaishuai\_guo@sdu.edu.cn}
\equalcont{These authors contributed equally to this work.}

\author*[2]{\fnm{Jianbo} \sur{Wang}}\email{wangjianbo@qiluhospital.com}
\equalcont{These authors contributed equally to this work.}

\affil[1]{\orgdiv{School of Control Science and Engineering}, \orgname{Shandong University}, \orgaddress{\city{Jinan}, \postcode{250061}, \country{China}}}

\affil[2]{\orgdiv{Department of Radiation Oncology}, \orgname{Qilu Hospital of Shandong University}, \orgaddress{\city{Jinan}, \postcode{250012}, \country{China}}}

\affil[3]{\orgdiv{Department of Civil \& Systems Engineering}, \orgname{Johns Hopkins University}, \orgaddress{\city{Baltimore}, \postcode{21218}, \country{United States}}}

\affil[4]{\orgname{Cognicore Artificial Intelligence Co., Ltd.}, \orgaddress{\city{Jinan}, \postcode{250100}, \country{China}}}


\abstract{With the rapid development of large language models (LLMs), their application to cell type annotation has drawn increasing attention. However, general-purpose LLMs often face limitations in this specific task due to the lack of guidance from external domain knowledge. To enable more accurate and fully automated cell type annotation, we develop a globally connected knowledge graph comprising $18850$ biological information nodes, including cell types, gene markers, features, and other related entities, along with 48,944 edges connecting these nodes, which is used by LLMs to retrieve entities associated with differential genes for cell reconstruction. Additionally, a multi-task reasoning workflow is designed to optimise the annotation process. Compared to general-purpose LLMs, our method improves human evaluation scores by up to $0.21$ and semantic similarity by $6.1$\% across multiple tissue types, while more closely aligning with the cognitive logic of manual annotation. Meanwhile, it narrows the performance gap between large and small LLMs in cell type annotation, offering a paradigm for structured knowledge integration and reasoning in bioinformatics.}


\maketitle
In single-cell RNA sequencing analysis, cell type annotation refers to the task of assigning each cell or cluster of cells to a biologically meaningful identity (e.g., T cells, B cells, epithelial cells) based on its gene expression profile. It is essential for downstream analyses such as tissue composition estimation, disease subtype discovery, and developmental trajectory reconstruction ~\cite{zhang2019cellassign, miao2020scaf}. Achieving precise cell type annotation through manual labeling typically requires two key steps: annotators retrieve the relevant top differentially expressed marker genes  \cite{meng2023singlecellbase, cellmarker, franzen2019panglaodb, patil2022cellkb} and integrate this information with their domain expertise to make informed decisions. However, this process is often complicated by factors such as overlapping marker genes, cellular heterogeneity, and inconsistencies across tissue contexts \cite{yuan2017singlecell, lahnemann2020challenges}. Although various automated approaches have been developed, fully automated and precise cell type annotation remains a significant challenge.

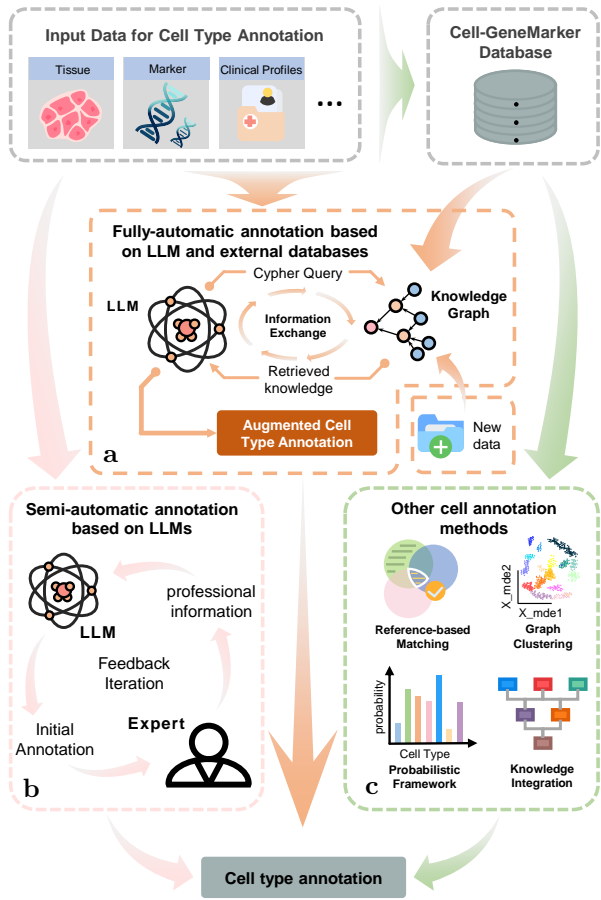
\begin{figure}[t]
\centering
\begin{tikzpicture}
    \node[anchor=south west, inner sep=0] (image) at (0,0) {\includegraphics[width=0.488\textwidth]{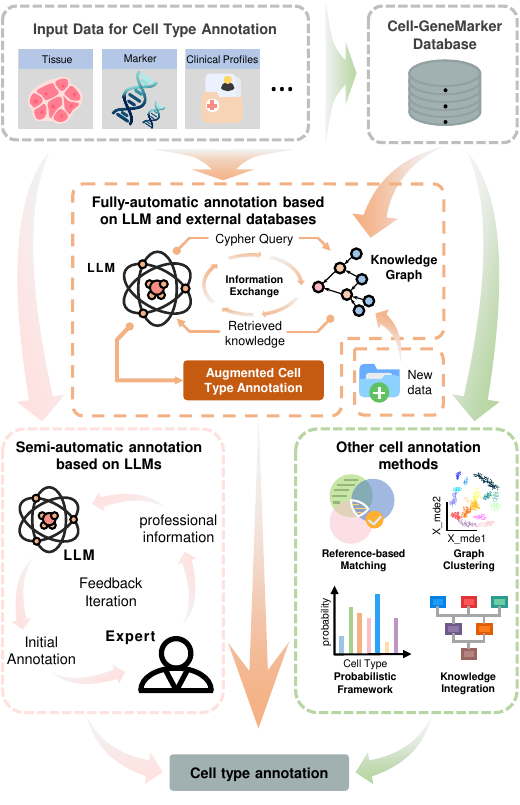}};
    \begin{scope}[xshift=0.02\textwidth, yshift=0.105\textwidth]
        \node {\textbf{b}}; \label{fig.1a}
    \end{scope}
    \begin{scope}[xshift=0.3\textwidth, yshift=0.105\textwidth]
        \node {\textbf{c}}; \label{fig.1b}
    \end{scope}
    \begin{scope}[xshift=0.085\textwidth, yshift=0.38\textwidth]
        \node {\textbf{a}}; \label{fig.1c}
    \end{scope}
\end{tikzpicture}
\caption{\textbar{\textbf{ Overview of cell type annotation methods.}}
          \textbf{a} Knowledge graph-driven LLM for automated annotation.
          \textbf{b} Semi-automated annotation with expert-LLM collaboration.
          \textbf{c} Traditional annotation methods prior to the advent of LLM.}\label{fig.1}
\end{figure}

Prior to the emergence of Transformer-based \cite{transformer} large language models (LLMs) \cite{gpt3, GPT4RS}, traditional approaches (Fig.~\hyperref[fig.1c]{1c}), such as clustering analysis and feature matching, were applied  for automated cell type annotation ~\cite{ScType, SingleR, CLForm}. These tools typically depend on predesigned pipelines or models trained on specific datasets. While effective in many routine scenarios, these methods often struggle to generalise beyond the distribution of their reference data or training data \cite{ye2024reliability, pasquini2021autoannotation}. Additionally, they offer limited support for user-specific objectives that require flexible or customised reasoning.

With the advancement of LLMs, their potential for fully or semi-automated cell annotation has also been explored (Fig.~\hyperref[fig.1b]{1b}). These models are typically pretrained on large-scale corpora, including relevant genomic and molecular biology databases, enabling them to capture diverse biological semantics and gene-function associations and achieve more accurate annotation results than traditional methods \cite{celltypeGPT}.

However, general-purpose LLMs, not optimised for specific downstream tasks during pretraining, exhibit suboptimal performance in specialised domains \cite{zheng2024finetuning, chen2024docoa}. Existing approaches have attempted to fine-tune LLMs for cell type annotation \cite{cell2sentence}, but such methods often fail to fully capture the intricate relationships between genes and cell types, while continuous fine-tuning can result in catastrophic forgetting of previously learned data \cite{luo2025empirical}. 

Graph Retrieval-Augmented Generation (GraphRAG), an enhancement technique for LLMs \cite{GraphRAG}, enables LLMs to extract entities and their relationships from unstructured text. It constructs a structured knowledge graph and utilizes retrieval-Augmented mechanisms \cite{lewis2020rag} to enhance its comprehension of entity relationships. In the medical domain, this method has been proposed \cite{KG-LLM-medicine-idea}, and its effectiveness has been demonstrated \cite{KG-LLM-medicine-work1, KG-LLM-medicine-work2, KG-LLM-medicine-work3}. In biological research, although existing studies have employed external data to enhance LLMs' biological insights \cite{DrBioRight}, the construction of knowledge graphs has not been utilized to enhance LLMs' retrieval and comprehension of biological entity information, and most of these studies have focused on relatively broad tasks such as medical and biological question answering, rather than being optimised for concrete, domain-specific tasks.

A potential limitation contributing to this phenomenon may stem from the reliance of retrieval-augmented generation methods on high-quality, task-specific datasets \cite{barnett2024ragfailures}. In particular, GraphRAG requires clearly defined entity structures and contextual associations to support effective querying and reasoning by large language models \cite{GraphRAG}. However, many existing biomedical data resources either lack structured format or do not provide semantic hierarchies tailored to specific tasks \cite{berlanga2012semantic, livingston2015kabob}. Recent studies have attempted to leverage large language models to collect information and extract entities from biomedical literature in order to complement or refine existing databases, using the extracted information as an auxiliary reference for downstream tasks \cite{extraction-work1, extraction-work2, extraction-work3}, but how to integrate this information with LLMs to enhance their reasoning abilities in downstream tasks has not yet been explored.

To address these challenges, we propose equipping LLMs with a refined knowledge graph designed for cell type annotation, integrating knowledge graph-based retrieval-augmented generation into the reasoning chain of cell type annotation. We design a retrieval-augmented mechanism that mirrors the human annotation process, in which experts search databases for relevant information and use it to make cell type decisions. The LLM follows a similar workflow by retrieving and reasoning over the knowledge graph to support automated annotation, thereby enhancing both the accuracy and interpretability of its predictions (Fig.~\hyperref[fig.1a]{1a}).


Specifically, to enhance interpretability and flexibility, we designed a modular, multi-task workflow that decomposes cell type annotation into subtasks such as broad cell type retrieval, marker–feature selection, and final decision making. The entire process is conceptualised as a form of cell name reconstruction, which we refer to as ReCellTy ("Reconstructing Cell Types"). In this framework, the LLM generates Cypher queries to interact with the structured knowledge graph, retrieving broad cell types and relevant features associated with differentially expressed marker genes. These intermediate outputs are subsequently processed by specialised agents that filter, organise, and reason over the retrieved information. The final agent integrates these results to determine the most probable cell type label (Fig.~\hyperref[fig.2c]{2c}). This structured design improves annotation consistency and enables the detection of rare or underrepresented cell types that are often missed by conventional approaches.

\section*{Results}
\subsection*{Data processing}
\begin{figure*}[t]
\centering
\begin{tikzpicture}
    \node[anchor=south west, inner sep=0] (image) at (0,0) {\includegraphics[width=1\textwidth]{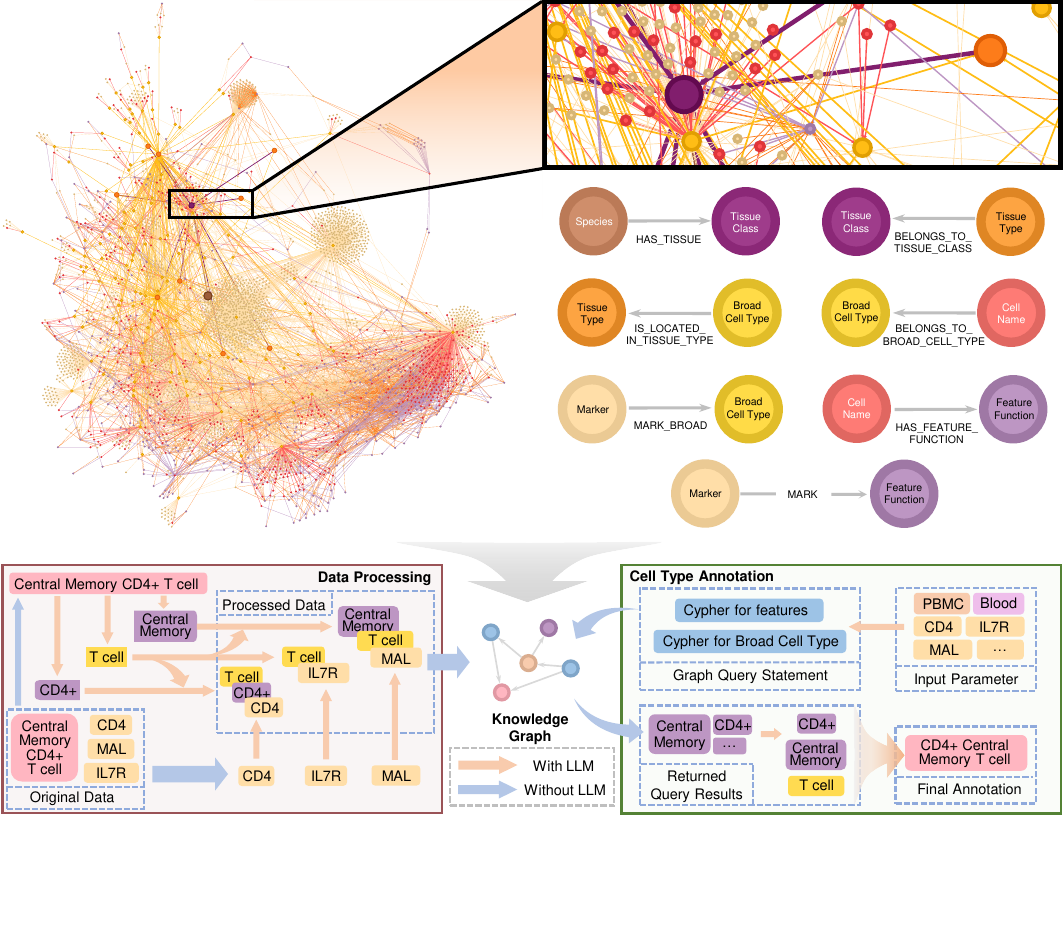}};
    \begin{scope}[xshift=0.01\textwidth, yshift=0.765\textwidth]
        \node {\textbf{a}}; \label{fig.2a}
    \end{scope}
    \begin{scope}[xshift=0.522\textwidth, yshift=0.6\textwidth]
        \node {\textbf{b}}; \label{fig.2b}
    \end{scope}
    \begin{scope}[xshift=0.01\textwidth, yshift=0.26\textwidth]
        \node {\textbf{c}}; \label{fig.2c}
    \end{scope}
\end{tikzpicture}
\caption{\textbar{\textbf{ Structure and method of the knowledge graph-based cell type annotation                               framework.}}
          \textbf{a} Visual representation of a subset of the data within the graph database.
          \textbf{b} Seven node types and their relationship types within the graph database.
          \textbf{c} Data processing pipeline and cell annotation question-answering            workflow.}\label{fig.2}
\end{figure*}

We initially attempted to construct a structured knowledge graph directly from the original CellMarker2.0 database \cite{cellmarker}, which provides curated associations between marker genes and specific cell types across various tissues. However, in practical applications, we found that LLMs struggled to accurately determine the most appropriate cell type from a large pool of candidates based solely on top differentially expressed markers. This difficulty stems from the complex and often overlapping relationships between marker genes and cell types. To address this issue, we optimised the raw CellMarker2.0 data by converting the structured marker–cell mappings into unstructured textual descriptions. We then prompted the LLM to extract richer associations from these texts, including marker to broad cell type links, marker to cell feature relationships, and higher-level biological context that is not explicitly represented in the original CellMarker2.0 database (Fig.~\hyperref[fig.2c]{2c}).

To systematically parse the associations between marker genes and cell features, we developed an end-to-end data processing pipeline based on the raw CellMarker2.0 database files (Cell\_marker\_Human.xlsx and Cell\_marker\_Seq.xlsx) \cite{cellmarker}. Initially, raw data entries were categorised according to the ‘tissue\_class’ field, generating tissue-specific subsets. Within each subset, cells were further refined across four attribute dimensions, ‘cell\_name’, ‘cell\_type’, ‘cancer\_type’, and ‘tissue\_type’, to establish a hierarchical structure, where each distinct cell type was assigned a dedicated CSV (Comma-Separated Values) file. This multi-level decomposition strategy effectively isolates irrelevant tissues and cells, significantly enhancing the LLM’s analytical ability to focus on feature-marker relationships for a single cell type.

For each generated CSV file, we designed a structured prompt template that takes cell name and the full marker gene list (‘marker’ header and ‘symbol’ header) as input. The LLM was tasked with three core functions: parsing function and feature descriptions from cell names, identifying broad cell type classifications, and establishing explicit feature-marker mappings. The model outputs CSV-formatted data containing these content, which we extracted from the text responses and horizontally merged with additional raw data from the original files, forming an enhanced cell feature-marker data. For example, in the case of Central Memory CD4+ T cells, the marker gene CD4 is directly linked to the naming feature CD4+, and since CD4 is biologically associated with T cells, we further mapped CD4 to T cell with CD4+ feature, as also the function Central Memory. This pipeline systematically covered all raw records for human species in CellMarker2.0, ensuring the completeness and traceability of feature association relationships.

Finally, the processed data, with cell types as the smallest unit, were first vertically merged by tissue to form individual datasets for each tissue, and then all tissue datasets were combined into a unified feature-marker association CSV database, which was then converted into a graph database to construct a global knowledge graph. Although this database was primarily designed to assist LLMs in retrieval and reasoning tasks, it remains accessible for human users to query related gene markers, cell types, and features. Therefore, we also developed an Excel spreadsheet to facilitate more intuitive data visualization (Supplementary Table 1).

Using this approach, we structurally reconstructed two datasets from the CellMarker2.0 database labeled as `Human', encompassing over $78,000$ raw entries. The final dataset consists of $1,528$ cell-naming feature and function components, $959$ broad cell types, $12,429$ gene markers and $61,049$ feature-gene-cell type associations.

Notably, by leveraging the knowledge embedded within LLMs, we automated the data processing flow, eliminating the reliance on manual analysis. This approach transforms traditionally time-consuming and inefficient manual tasks into a scalable, high-throughput data processing pipeline. Furthermore, our method is adaptable to new datasets: By inputting gene and cell type information, the LLM-based framework rapidly parses, extracts biological data, and stores them in a structured format that is directly compatible with the knowledge graph, which significantly accelerates the integration of biological knowledge.

\subsection*{Feature-marker GraphRAG}
Based on the restructured dataset, we constructed a structured knowledge graph by defining biological entity nodes and their relationship types, which were then stored in a graph database to facilitate efficient retrieval. This graph encompasses 18,850 biological entity nodes, including genes, cell types, and their associated features and functions, etc, as well as 48,944 connecting relationships between these nodes. The highly intricate structure tightly integrates various types of biological information, enabling the LLMs to uncover biological associations and interactions akin to human-level insights. We also visualised a subset of the database to provide an intuitive representation of the graph structure (Fig.\hyperref[fig.2a]{2a}).

Building on feature-marker association CSV data, we constructed a structured knowledge graph using the Neo4j graph database \url{https://neo4j.com/}. Specifically, we developed a Cypher query code to sequentially convert preprocessed CSV data into graph entities.In this Cypher query, we defined seven core node types, such as Marker (gene marker) and FeatureFunction (features or functions in cell name), and established seven types of semantic relationships, such as MARK (annotation from Marker to FeatureFunction) and HAS\_FEATURE\_FUNCTION (annotation from CellName to FeatureFunction) (Fig.\hyperref[fig.2b]{2b}), forming a multi-layered cellular feature network.

The construction of the knowledge graph integrates a large number of biological entities and their direct or indirect relationships. By defining extensive biological entity nodes and the relationships among them, it systematically captures multi-level cellular information. Moreover, this extensive and densely connected structure allows complex biological associations to manifest across different cellular and tissue levels, enabling large language models to more effectively capture these intricate interactions and thereby enhance their overall reasoning capabilities. Consequently, the knowledge graph transcends localized or isolated datasets, forming a global data network.

For retrieval, we leveraged the GraphCypherQAChain module from the LangChain \url{https://www.langchain.com}, integrating retrieval and question-answering into a high-efficiency querying system. The core mechanism involves feeding the LLM a schema description of the knowledge graph, reinforcing its understanding of the knowledge graph’s topology and improving the quality of Cypher query generation. The query workflow proceeds as follows: first, the LLM parses the user’s natural language input to determine key retrieval elements. Second, a Cypher query is dynamically generated based on the knowledge graph’s structural features. Third, the query is executed to retrieve data from the knowledge graph. Finally, the retrieval results are combined with the original query to generate a formatted response.

Specific to cell type annotation, we designed a dual-retrieval augmentation mechanism. Each retrieval augmentation includes a retrieval module and a simple question-answering module. One retrieval augmentation gets broad cell types associated with the given markers, while the other gets features and functions related to the markers. For the retrieval modules, we constructed two distinct Cypher query templates and embedded them into the retrieval prompts as references for the LLMs, aiming to improve the accuracy of generated retrieval code and the usability of the returned entity node formats. For the question-answering modules, we require the LLMs to return a text block in the form of “Marker1: Broad cell type1, Broad cell type2” or “Marker2: Feature1/Function1, Feature2/Function2” based on the retrieved broad cell types and features/functions. We then extract this text block from the responses as the standardised output. The standardised outputs from these two retrieval augmentation serve as the input evidence for the downstream reasoning system.


\subsection*{Multi-task reasoning framework}
To enable human-like decision logic in cell type annotation, we developed a modular workflow system built on top of the knowledge graph retrieval architecture (Broad CellType Query Task and Feature/Function Query Task). This workflow, implemented by designing tailored prompts to leverage LLMs (all prompts, including data processing task, are provided in Supplementary Prompt), integrates the above-mentioned dual-retrieval augmentation mechanism and employs a multi-task system to refine knowledge-driven reasoning. Specifically, the workflow consists of the following components:

\textbf{Broad CellType Query Task} processes the input list of top differentially expressed gene markers to retrieve their associated broad cell types from the structured graph dataset. This task involves querying the knowledge graph to identify and extract relevant broad cell type entities connected to the input markers. The retrieved results are then consolidated and formatted into a standardized Marker-CellType correspondence table, providing a clear mapping between each marker gene and its possible broad cell types.

\textbf{Broad CellType Selection Task} takes the summarized Marker-CellType correspondence information generated by the Broad CellType Query Task and performs a decision-making process to determine the most probable broad cell type. This selection step analyzes the retrieved query data and prioritizes the broad cell types that best represent the input gene set, thereby narrowing down the classification for subsequent annotation stages.

\textbf{Feature/Function Query Task} operates similarly to the Broad CellType Query Task but focuses on retrieving named cellular features or functions associated with the input markers. Through querying the graph dataset, this task extracts detailed feature or function entities linked to each marker gene, enabling the characterization of marker-specific biological attributes beyond broad cell type classification.

\textbf{Feature/Function Selection Task} refines the results from the Feature/Function Query Task by filtering the marker–feature mappings to select a subset of 2 to 3 features with the highest confidence. This selective process reduces the complexity of the decision space for downstream modules, ensuring that only the most relevant and informative features are considered in the final annotation pipeline.

\textbf{CellType Annotation Task} integrates the multiple layers of information collected from prior tasks, including the predicted broad cell type, the selected cellular features/functions, and the original set of marker genes. By combining these diverse data sources, this task produces the final cell type annotation label, enabling comprehensive and accurate characterization of cell identity.

To illustrate the retrieval and application workflow of GraphRAG combined with multi-task reasoning, we present a cell type annotation example supported by knowledge graph visualization.

In this example, the input consists of a set of differentially expressed gene markers, 'IL7R, TMSB10, CD4, ITGB1, LTB, TRAC, AQP3, LDHB, IL32, MAL', along with the specified tissue context, Blood and Peripheral blood. The ground-truth annotation for this example is CD4+ Central Memory T cell.

\begin{figure*}[t]
\centering
\begin{tikzpicture}
    \node[anchor=south west, inner sep=0] (image) at (0,0) {\includegraphics[width=1\textwidth]{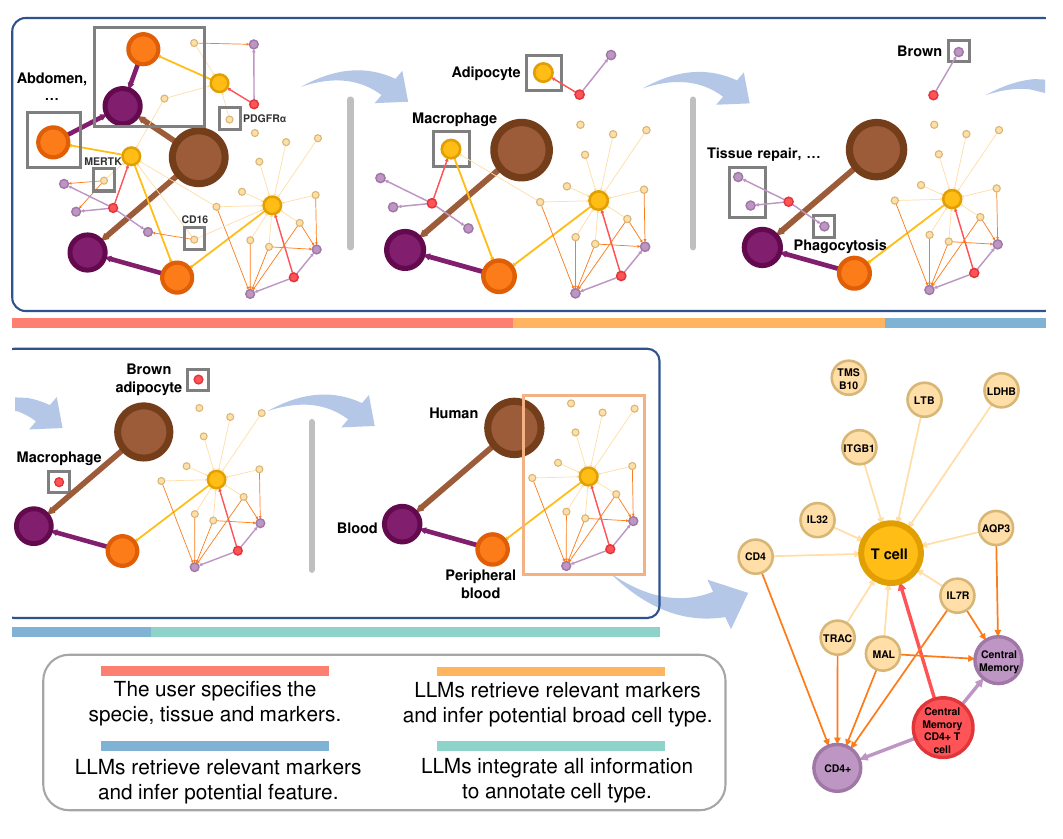}};
    \begin{scope}[xshift=0.04\textwidth, yshift=0.75\textwidth]
        \node {\textbf{a}}; \label{fig.3a}
    \end{scope}
    \begin{scope}[xshift=0.35\textwidth, yshift=0.75\textwidth]
        \node {\textbf{b}}; \label{fig.3b}
    \end{scope}
    \begin{scope}[xshift=0.67\textwidth, yshift=0.75\textwidth]
        \node {\textbf{c}}; \label{fig.3c}
    \end{scope}
    \begin{scope}[xshift=0.04\textwidth, yshift=0.443\textwidth]
        \node {\textbf{d}}; \label{fig.3d}
    \end{scope}
    \begin{scope}[xshift=0.35\textwidth, yshift=0.443\textwidth]
        \node {\textbf{e}}; \label{fig.3e}
    \end{scope}
    \begin{scope}[xshift=0.7\textwidth, yshift=0.443\textwidth]
        \node {\textbf{f}}; \label{fig.3f}
    \end{scope}
\end{tikzpicture}
\caption{\textbar\textbf{ A visualization workflow of GraphRAG illustrating its integration with multi-task reasoning for cell type annotation. The figure shows how graph-based retrieval and reasoning modules collaborate to extract and synthesize biological information for accurate cell type identification.}
          \textbf{a} The original global knowledge graph, containing both relevant information for this annotation and noise nodes.
          \textbf{b} The knowledge graph filtered by user input, removing irrelevant tissues and gene markers.
          \textbf{c} The knowledge graph after LLMs determine broad cell type related to markers, with unrelated broad cell types removed.
          \textbf{d} The knowledge graph after LLMs determine features and functions related to markers, with unrelated features and functions removed.
          \textbf{e} The knowledge graph resulting from the LLM’s correct cell type annotation, integrating information on tissues, markers, broad cell types, and features.
          \textbf{f} The knowledge graph information of the four node types most relevant to this annotation process.
          }\label{fig.3}
\end{figure*}

To simulate a realistic and complex knowledge environment, we introduce noise by incorporating unrelated nodes into the visualized meaningful subgraph, representing the structure of the global knowledge graph (Fig.\hyperref[fig.3a]{3a}). By progressively reducing task-irrelevant information in the visualised graph, we reconstruct the retrieval and reasoning process of our multi-task workflow.

Initially, after the user specifies the species, tissue, and input markers, the knowledge graph performs a coarse filtering step, where potentially relevant entities are retained, while unrelated tissue and marker nodes are removed (Fig.\hyperref[fig.3b]{3b}). Subsequently, the Broad CellType Query Task and Broad CellType Selection Task are performed. The LLMs retrieve and evaluate broad cell types associated with the input markers. During this step, unrelated broad cell type nodes are pruned from the visual graph, preserving only the most likely broad cell type node (Fig.\hyperref[fig.3c]{3c}). In the next stage, the Feature/Function Query Task and Feature/Function Selection Task are executed. Here, LLMs infer key features and functions linked to the input markers, further refining the information landscape. The visual representation is updated accordingly by removing unrelated nodes (Fig.\hyperref[fig.3d]{3d}). Finally, the CellType Annotation Task integrates all previously retrieved components—predicted broad cell types, selected features/functions, and original markers—to generate the final cell type annotation. At this point, only the node representing CD4+ Central Memory T cell remains in the visualized subgraph (Fig.\hyperref[fig.3e]{3e}).

From this process, we can isolate four types of graph nodes most directly involved in the annotation task: marker genes, broad cell types, functional/feature components, and final cell names (Fig.\hyperref[fig.3f]{3f}). The relationships among these node types highlight how multi-step reasoning over structured biological knowledge supports accurate, human-like cell type annotation.

\subsection*{Evaluation scores}
To demonstrate the effectiveness of our method, we conducted tests on 11 tissue types from the Azimuth dataset using four non-reasoning large language models. Given that the Claude 3.7 Sonnet model supports both standard and extended reasoning modes, we uniformly adopted the standard mode for all experiments in this study. For reproducibility, we performed five independent question-answering iterations for each row of differential genes, and the most frequent result was taken as our experimental evaluation benchmark (Supplementary Table 2).

Regarding the scoring mechanism design, we employed both manual evaluation and semantic evaluation scoring strategies to compare the annotation results of ReCellTy with the manual annotations in the dataset. Since our method supports the visual display of relevant information retrieved from the graph dataset, we introduced an intermediate process adjustment during manual evaluation to refine the accuracy of the final cell type scores. Semantic evaluation was assessed using an embedding model, where text annotations were converted into vectors (Supplementary Data), and the cosine similarity between them was calculated.

\begin{figure*}[t]
\centering
\begin{tikzpicture}
    \node[anchor=south west, inner sep=0] (image) at (0,0) {\includegraphics[width=1\textwidth]{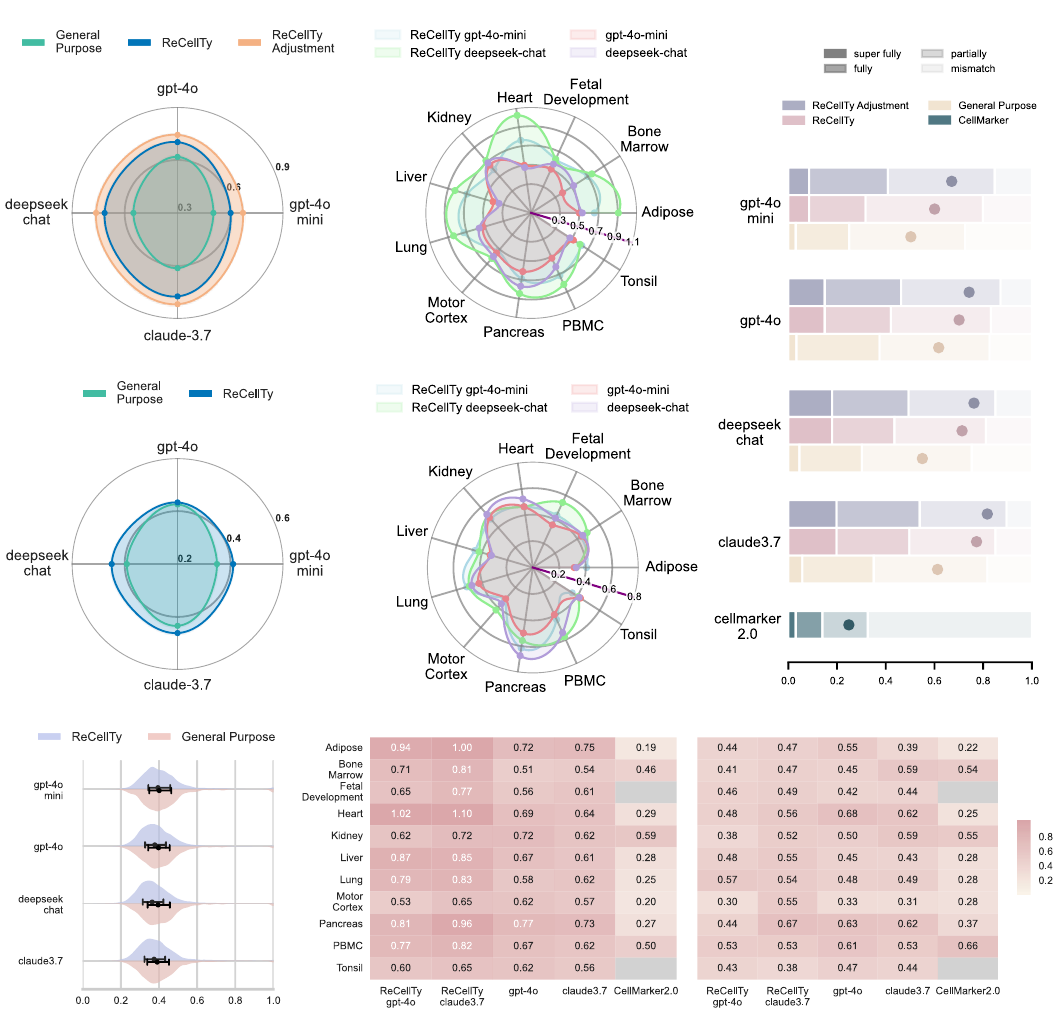}};
    \begin{scope}[xshift=0.01\textwidth, yshift=0.955\textwidth]
        \node {\textbf{a}}; \label{fig.4a}
    \end{scope}
    \begin{scope}[xshift=0.33\textwidth, yshift=0.955\textwidth]
        \node {\textbf{b}}; \label{fig.4b}
    \end{scope}
    \begin{scope}[xshift=0.7\textwidth, yshift=0.925\textwidth]
        \node {\textbf{c}}; \label{fig.4c}
    \end{scope}
    \begin{scope}[xshift=0.01\textwidth, yshift=0.615\textwidth]
        \node {\textbf{d}}; \label{fig.4d}
    \end{scope}
    \begin{scope}[xshift=0.33\textwidth, yshift=0.615\textwidth]
        \node {\textbf{e}}; \label{fig.4e}
    \end{scope}
    \begin{scope}[xshift=0.01\textwidth, yshift=0.29\textwidth]
        \node {\textbf{f}}; \label{fig.4f}
    \end{scope}
    \begin{scope}[xshift=0.3\textwidth, yshift=0.29\textwidth]
        \node {\textbf{g}}; \label{fig.4g}
    \end{scope}
    \begin{scope}[xshift=0.66\textwidth, yshift=0.29\textwidth]
        \node {\textbf{h}}; \label{fig.4h}
    \end{scope}
\end{tikzpicture}
\caption{\textbar{\textbf{ Performance evaluation.}}
          \textbf{a} Human evaluation scores of four large models under different methods.
          \textbf{b} Human evaluation scores of two large models across various tissues.
          \textbf{c} Overall human evaluation scores for each method.
          \textbf{d} Semantic evaluation scores of four large models under different methods.
          \textbf{e} Semantic evaluation scores of two large models across various tissues.
          \textbf{f} Intra-group semantic variance of annotation results for each model and method.
          \textbf{g} Human evaluation scores of two models and CellMarker 2.0 across various tissues.
          \textbf{h} Semantic evaluation scores of two models and CellMarker 2.0 across various tissues.}\label{fig.4}
\end{figure*}

The experimental results demonstrate better performance of ReCellTy compared to general-purpose LLMs and the CellMarker 2.0 annotation across all four tested models and both scoring mechanisms (Figs.~\hyperref[fig.4a]{4a} and~\hyperref[fig.4d]{4d}). After process adjustment, ReCellTy achieved an average score improvement of 0.18 across four models. For each tissue, ReCellTy achieved higher manual evaluation scores in the majority of tissues (Figs.~\hyperref[fig.4b]{4b} and~\hyperref[fig.4g]{4g}). For example, the DeepSeek-Chat exhibited a 0.55 increase on the Heart dataset and a 0.48 improvement on the Liver dataset.

In manual evaluation, ReCellTy effectively bridges the performance gap between smaller and larger models (Fig.~\hyperref[fig.4c]{4c}). The general-purpose GPT-4o-mini model achieved a human evaluation score of only 0.50, but the process adjustment ReCellTy boosted its performance to 0.67, surpassing larger models with more parameters and training data, including GPT-4o (0.62), DeepSeek-chat (0.55), and Claude 3.7 (0.61). This suggests that integrating an external graph-based knowledge base can effectively overcome the limitations of general-purpose LLMs when applied to specialised tasks. In particular, it helps narrow the performance gap caused by differences in architecture, parameter scale, data volume, and training objectives between task-specific and general-purpose models.

To investigate the diversity of cell types generated by ReCellTy, we calculated the intra-group semantic similarity score for each model and method (Fig.~\hyperref[fig.4f]{4f}). The results showed that ReCellTy exhibited lower similarity compared to general-purpose large language models. This lower similarity indicates higher diversity, suggesting that by integrating knowledge graph information and employing retrieval-augmented generation, ReCellTy can focus on a more diverse range of cell types, significantly enhancing the diversity of annotation results compared to general-purpose LLMs.

In practical use, we observed that leveraging an external graph database for retrieval and reasoning offers significant advantages in reducing hallucinations in large language models. A high-quality knowledge graph can accurately associate top differentially expressed genes with corresponding broad cell types and functional features, thereby reducing mismatches and reasoning errors that may arise from the model’s autonomous judgment. Moreover, when the LLM fails to retrieve relevant information from the graph in a given subtask, it explicitly responds with "I don't know the answer." This mechanism supports more rational evaluation and critical interpretation of the model's outputs by the user.

\section*{Discussion}
\noindent
In this work, we construct a structured biological knowledge graph and develop a modular multi-task reasoning framework for automated cell type annotation, with a particular emphasis on leveraging retrieval-augmented LLMs. The system integrates knowledge retrieval with prompt-based reasoning to enable accurate annotation. Based on this methodology, we also developed a practical annotation tool and user interface to support interactive cell type annotation (see Materials and Methods). Experimental results show that our framework achieves strong annotation performance and improves model interpretability. By tracing the intermediate outputs of the reasoning process, such as selected broad cell type and marker–feature associations, we provide a reference example for integrating structured knowledge and LLM-based reasoning in complex biological annotation tasks.

For knowledge graph construction, we introduced LLM-based information extraction to analyse and restructure biological data from CellMarker2.0. Instead of relying solely on the original marker-to-cell type mappings, we converted these structured entries into unstructured natural language prompts, enabling LLMs to extract deeper semantic relationships, such as marker–feature, marker–broad cell type, and feature–cell associations. The resulting outputs were standardised and stored in a graph database, which supports efficient, schema-based Cypher querying. This approach allows the knowledge graph to serve not only as a passive data repository but as an active reasoning substrate, supporting both analysis and fully automated annotation pipelines. By systematically defining entities and their relationships, the graph captures multi-level biological information across tissues and cell types.

To enable interpretable and fine-grained reasoning, we designed a modular multi-task workflow that decomposes the cell type annotation process into sequential subtasks: broad cell type retrieval, broad cell type selection, feature/function retrieval, feature/function selection, and final cell type annotation. These tasks are implemented by designing prompts that enable LLMs to interact effectively with the knowledge graph, using Cypher query for retrieval and formatted response blocks as standardised outputs. This design enables the system to progressively filter and refine candidate annotations, emulating expert decision logic. The multi-step structure not only improves annotation accuracy but also supports transparency, as each decision point can be traced and analysed independently. Furthermore, by separating reasoning into subtasks, the workflow remains flexible and extensible to different LLMs and datasets, improving its adaptability and robustness in real applications.

While the progress achieved in this task highlights the practicality of our framework, some challenges still need to be addressed. For example, although the LLM-powered automated parsing in the data processing stage greatly enhances efficiency, it may occasionally produce biologically suboptimal feature–marker associations due to the inherent uncertainty in generative outputs. Moreover, the serial structure of the multi-task workflow involves multiple rounds of retrieval and question answering, which increases token usage and leads to higher computational costs for each annotation request. for knowledge graph, our current graph only establishes links between biologically related nodes, without capturing the finer-grained semantics of relationships between different biological entities. More precise modelling of these inter-node relationships could further improve the expressiveness of the graph and enhance the interpretability of the reasoning process.

In addition, our semantic evaluation strategy may underestimate model performance in cases where more specific and informative annotations are produced. Although semantic evaluation shows an overall improvement in scores, we observed that it may assign lower scores to more specific cell type annotations. For example, in human evaluation, we assign a score of 1.5 when the model gives a more specific subtype, result shows that GPT-4o with ReCellTy reaches a score of 1.02, achieving a score improvement of 0.33 on the Heart dataset. This demonstrates that the model successfully annotated more specific cell subtypes. However, the semantic evaluation showed a 20\% decrease. Although ReCellTy demonstrated overall improvement in semantic evaluation (average improvement: 3.8\%; GPT-4o-mini: up to 6.1\%), its ability to annotate more specific cell types, which should have been an advantage over other methods, instead resulted in less noticeable improvements in semantic scores at the level of individual tissues (Figs.~\hyperref[fig.4e]{4e} and~\hyperref[fig.4h]{4h}).

Looking ahead, the continued advancement of large-scale initiatives such as the Human Genome Project (HGP) \cite{hood2013hgp} and the Human Cell Atlas (HCA) \cite{regev2018hca, rozenblattrosen2017hca} is expected to facilitate the development of unified and high-quality dynamic cell databases. Within this evolving technological landscape, future research could explore leveraging LLM agents as both managers and users of such databases. On one hand, enabling real-time updates and maintenance of the graph dataset through autonomous agents, and on the other, designing human-like multi-agent collaborative workflows tailored to various cell-gene tasks. This database-empowered architecture has the potential to significantly enhance the practical utility of LLMs in the field of cellular genomics.

\section*{Methods}
\subsection*{Comparative Methods}
The method proposed in this study achieves enhanced cell type annotation performance by integrating and optimizing LLMs with the knowledge augmentation strategy of the CellMarker 2.0 database. To validate the effectiveness of our method, we selected the following two methods for comparative experiments:

\textbf{CelltypeGPT.} This is an LLM-based cell annotation tool \cite{celltypeGPT}. Since ReCellTy employs a single-row differentially expressed gene processing mode, to ensure fairness in the comparison, we adjusted the prompting strategy of CelltypeGPT and denote its method as 'general-purpose'. The adjusted prompt template is as follows:

'Identify cell types of {TissueName} cells using the following markers. Only provide the cell type name. Do not show numbers before the name. Some can be a mixture of multiple cell types. {GeneList}'

TissueName and GeneList will be replaced with the actual tissue and differentially expressed gene list, respectively.

\textbf{CellMarker2.0.} The official tool, directly accessing the webpage interface of this database for annotation.

\subsection*{Evaluations}
We used two types of evaluation to assess annotation performance: manual evaluation based on predefined scoring rules that reflect human interpretation of annotation correctness, and semantic evaluation that computes cosine similarity between annotation labels using embedding models. The combination of human judgment and embedding-based comparison allows us to evaluate both annotation accuracy and semantic consistency from different perspectives.

\textbf{Manual evaluation.} For manual evaluation, we improved upon the evaluation framework of CelltypeGPT. The criteria for determining the final annotation results are as follows: if the automatically annotated result is more specific than the manual annotation, it is defined as "super fully"; if the two are exactly the same, it is "fully"; if the two belong to the same major cell type or have a differentiation-related connection, it is defined as "partially"; and completely unrelated is "mismatch". In the intermediate process evaluation, if the major cell type and the selected relevant features can be combined to produce a cell type that is exactly the same as the manual annotation, it is "fully"; partially related is "partially"; and completely unrelated is "mismatch". The four evaluation levels (super fully/fully/partially/mismatch) are assigned weights of 1.5, 1.0, 0.5, and 0, respectively, and then the average score within the group is calculated.

\textbf{Semantic evaluation.} For semantic similarity, we used OpenAI's text-embedding-3-small to encode the cell annotation results and then calculated the cosine similarity between the vector representations of the manual and automatic annotations. Subsequently, we normalized the cosine similarity within the group and divided it into five equal intervals from high to low, assigning scores of 1, 0.75, 0.5, 0.25, and 0, respectively, and then calculated the average score within the group.

\subsection*{Application}
To enhance system interpretability and assist user decision-making, we developed a multi-stage visualisation system and an interactive user interface (UI) tailored for cell type annotation tasks. This UI is designed to facilitate transparency, interactivity, and traceability across the entire annotation process. It is composed of the following key modules:

\textbf{Input Layer}, which allows users to submit top differentially expressed marker genes and select the tissue. The option to specify tissue type serves to constrain the query space, thereby improving retrieval specificity and annotation precision. To address data scarcity in certain tissues, we also provide a global query mode that removes tissue-specific constraints, enabling broader access to the knowledge graph and maximising the utility of available biological information.

\textbf{Processing Layer}, which provides real-time feedback on system status, including the annotation progress of each retrieval module and the execution status of the multi-task workflow. This layer serves primarily as a monitoring interface to help users track the overall progress of the annotation process.

\textbf{Output Layer}, which displays the final cell type annotation along with the intermediate reasoning steps, such as selected broad cell types and functional features. This transparent output format enables users to trace the full decision-making path from input to prediction, supporting both interpretability and post hoc validation.

Additionally, to promote integration with widely used single-cell analysis pipelines such as Seurat, we also developed a Python-based package that processes input gene lists and interfaces with the annotation system. Although the package is implemented in Python, it could support interoperability with R-based workflows through cross-language tools such as "rpy2" (which allows calling R from Python) and "reticulate" (which allows calling Python from R). This flexibility allows researchers working in either R or Python environments to incorporate the annotation tools into their analysis workflows.

\section*{Data availability}
The datasets and models used in this study are available from the following sources. The CellMarker 2.0 dataset can be downloaded at \url{http://www.bio-bigdata.center/CellMarker_download.html}, and the Azimuth dataset is available at \url{https://azimuth.hubmapconsortium.org/}. The language models employed, including GPT-4o-mini, GPT-4o, and text-embedding-3-small, are accessible via the OpenAI API at \url{https://openai.com/api/}. DeepSeek-chat and Claude 3.7 were accessed through their respective APIs at \url{https://platform.deepseek.com/sign_in} and \url{https://www.anthropic.com/api}.

\section*{Code availability}
The ReCellTy package and UI, together with their source code and associated data, are publicly available at \url{https://github.com/SSG2019/ReCellTy}. The original code framework, data processing pipeline, and experimental test datasets have also been released at \url{https://github.com/SSG2019/ReCellTy-paper}.

\section*{Acknowledgements}
This work was supported by National Natural Science Foundation of China No. 82172742 and 82473353.
\section*{Author contributions}
Conceptualization: D.H., Y.J., S.G., and J.W. Framework implementation: D.H., Y.J., R.C., and W.H. Tool development: D.H. and W.H. Result analysis: Y.J. and D.H. Supervision: S.G. and J.W. Writing: D.H., Y.J., R.C., W.H., S.G., and J.W.



\end{document}